# Indy Autonomous Challenge - Autonomous Race Cars at the Handling Limits

Alexander Wischnewski[1,*], Maximilian Geisslinger[2], Johannes Betz[3], Tobias Betz[2], Felix Fent[2], Alexander Heilmeier[2], Leonhard Hermansdorfer[2], Thomas Herrmann[2], Sebastian Huch[2], Phillip Karle[2], Felix Nobis[2], Levent Ögretmen[1], Matthias Rowold[1], Florian Sauerbeck[2], Tim Stahl[2], Rainer Trauth[2], Markus Lienkamp[2], Boris Lohmann[1]

[1] Chair of Automatic Control, Technical University of Munich
[2] Chair of Automotive Technology, Technical University of Munich
[3] Real-Time and Embedded Systems Lab, University of Pennsylvania
* Corresponding author: alexander.wischnewski@tum.de

**Abstract.** Motorsport has always been an enabler for technological advancement, and the same applies to the autonomous driving industry. The team TUM Autonomous Motorsports will participate in the Indy Autonomous Challenge in October 2021 to benchmark its self-driving software-stack by racing one out of ten autonomous Dallara AV-21 racecars at the Indianapolis Motor Speedway. The first part of this paper explains the reasons for entering an autonomous vehicle race from an academic perspective: It allows focusing on several edge cases encountered by autonomous vehicles, such as challenging evasion maneuvers and unstructured scenarios. At the same time, it is inherently safe due to the motorsport related track safety precautions. It is therefore an ideal testing ground for the development of autonomous driving algorithms capable of mastering the most challenging and rare situations. In addition, we provide insight into our software development workflow and present our Hardware-in-the-Loop simulation setup. It is capable of running simulations of up to eight autonomous vehicles in real time. The second part of the paper gives a high-level overview of the software architecture and covers our development priorities in building a high-performance autonomous racing software: maximum sensor detection range, reliable handling of multi-vehicle situations, as well as reliable motion control under uncertainty.

## 1    Introduction

### 1.1    Motivation

Autonomous driving promises to be one of the next major revolutions within the mobility sector. The research on this topic goes back to the pioneering works of Dickmanns et al. [1] on a self-driving Mercedes 500 SEL in the EUREKA-project Prometheus III reaching around 130 kph on a highway, as well as the results achieved by Thrope et al.

at Carnegie Mellon University with the vehicle NAVLAB [2]. Following their promising results, the DARPA Grand and Urban Challenges in 2004, 2005, and 2007 paved the way for thousands of researchers to work on this challenging technology [3, 4].

In recent years, benchmarks of autonomous driving algorithms on racetracks started to gather the interest of several research groups [5, 6]. This application poses several interesting questions, such as motion planning and control at the physical limits of the vehicle, reliable and robust perception at high speeds, as well as multi-vehicle dynamic scenarios. Furthermore, motorsport has always been on the cutting edge of technology in the automotive field and gathered talent and resources by creating a competitive environment. Therefore, it is no surprise that several events and competitions emerged in this field, such as the Formula Student Driverless [7, 8], the F1/10th race series [9], and the Roborace competition [6]. While all of those are mainly focused around time-trials or two-vehicle scenarios, the Indy Autonomous Challenge [10] aims to be the first multi-vehicle wheel-to-wheel race in the world. This paper will discuss the motivation of the TUM Autonomous Motorsports team to engage in this competition, present the competition itself as well as analyze the key performance factors we identified to build a winning autonomous racing software stack.

The remainder of this section is going to present the state-of-the-art related to autonomous racing. The second section presents the advantages of autonomous racing for research purposes and our reasoning to enter the Indy Autonomous Challenge. It is concluded with an overview of our development approach and the Hardware-in-the-Loop simulation used for the competition. The third sections outlines the key technical challenges aligned with the autonomous multi-vehicle race. Finally, the paper is concluded in section four.

## 1.2 State of the art

One of the first fields of research interests for autonomous vehicle (AV) racing was determining the time-optimal racing line and subsequent execution of fast qualifying-like laps. This problem has a substantial similarity with algorithms used for parameter optimization and lap time optimization in classical motorsports. Depending on the available computational resources and the chosen model fidelity, the resulting optimization problem can be solved in a range from a few milliseconds up to several minutes [11-14]. Several model-based control algorithms based on the well-known two-degree of freedom structure have been proposed to drive these optimal lines with real-world prototypes. Noteworthy examples are the autonomous Audi TT build by the Stanford University [5] and the DevBot 2.0 operated by the TUM Autonomous Motorsport team from Munich [6, 11]. The localization of these prototypes has usually been done based on high-quality DGPS receivers. Other works have extended the functionality to provide redundant LIDAR-based localization [15, 16]. All of these works demonstrated single-lap performances close to skilled human drivers.

While the time-optimal racing line calculation requires detailed knowledge about the available friction, this information is not considered in designing the two-degree of freedom control structures. This deficiency is overcome by more advanced model predictive control algorithms [7, 17]. The basic concept behind this design is to solve a receding horizon optimal control problem using a vehicle dynamics model to predict future behavior. After applying the first control input, the optimization is performed again based on the updated measurements. This strategy leads to a controller which tracks the ideal raceline also in the presence of uncertainty. The choice of the vehicle dynamics models has to be made by balancing the increasing computational demands in contrast to improved accuracy of more complex models [18]. Another strategy to mitigate model uncertainties is the application of machine learning methods to improve the model accuracy while driving [19-21].

In addition to these vehicle dynamics challenges, motion planning becomes much more complex in dynamic multi-vehicle scenarios with only a loose set of rules. These scenarios are considered non-convex, meaning that there are multiple, distinct behavior options leading to different locally time-optimal motion plans [22]. An example of this is the choice between overtaking an opponent on the left or the right shown by Figure 1. This challenge is difficult to resolve with gradient-based optimization algorithms as they tend to converge to a local solution based on their initialization. An approach to solve this problem is applying graph-based optimization algorithms as they lead to globally optimal solutions as proposed by [23]. However, this work splits the task of path and velocity planning to match computation time requirements which leads to sub-optimal solutions when the scenarios become more challenging. A different approach, solving the combinatorial nature of the problem with a game-theoretic strategy, is combined with a sophisticated model predictive control for simultaneous re-optimization and feedback control in [24-26]. Another critical challenge in generating high-performance trajectories at the physical limits is guaranteeing the recursive feasibility of the motion planning problem. This translates into the requirement that the motion planning problem stays feasible in the presence of constraints such as tire capabilities or track limits. The research within this area has been concentrated around single-vehicle scenarios with online lap time optimization [27, 28]. The theoretically sound extension to multi-vehicle scenarios is an unsolved problem and usually circumvented by the careful tuning of cost functions and optimization problem design.

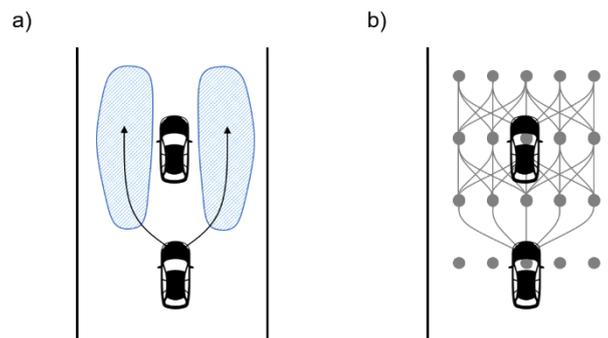

**Figure 1:** Overtaking maneuvers are non-convex optimization problems (a), where graph-based algorithms (b) still lead to globally optimal solutions by discretization in a spatio-temporal graph

Despite the challenges in motion planning and control, the reliable operation of an autonomous racecar requires a localization system that is ideally independent of GPS as this might not always be available. The research presented in [15] proposes an occupancy grid-based LIDAR-SLAM based on the AMCL algorithm and successfully demonstrates operation at 150 kph. The usage of Camera-SLAM is a promising alternative [29]; however, it has not yet been deployed in full-scale racing environments. Slower speed environments with specific features such as Formula Student Tracks with their cone markings have been tackled successfully using feature or landmark SLAM algorithms combined with separate detection pipelines [8, 30].

The remaining part of the software stack yet uncovered is the object detection algorithms generating reliable information about the opponents and their behavior on the racetrack. This is an inherent multi-vehicle situation characteristic and therefore the research in this area has not yet been as intense as on the motion planning and vehicle dynamics control part. Several applications of well-known general neural network architectures like YOLO have been proposed for landmark/feature detection in the Formula Student competition for cone detection [30]. Others used LIDAR clustering algorithms for this task [8]. The detection of opponent vehicles has been tackled by monocular depth estimation on real-world data by [31].

An approach completely different from the previous paragraphs' modular system is the application of deep neural networks to large parts of the autonomous driving tasks. The authors in [32] demonstrate good driving performance by proposing an approach where the images collected by a camera are mapped to a target trajectory to be followed by a low-level controller. In [33] a reinforcement learning algorithm was displayed, that learns competitive visual control policies through self-play in imagination. Although this method is not applied to real vehicles it provided interesting strategies for multi vehicle interaction.

## 2 Autonomous racing

### 2.1 Motorsport as a proving ground for autonomous vehicles

There are mainly two key challenges that need to be solved to bring autonomous vehicles on the streets: Firstly, the broader public will require AVs to perform at a superhuman safety level while being comfortable to drive. Even when software developers and engineers would develop such a system, the testing and safety certification would require autonomous vehicles to drive hundreds of millions of kilometers in common traffic scenarios to demonstrate their ability [34]. Secondly, a sound and widely accepted regulatory framework to deploy autonomous systems in the general public needs to be established [35]. However, we will not cover the latter one and instead focus on the technological aspects towards solving the first challenge.

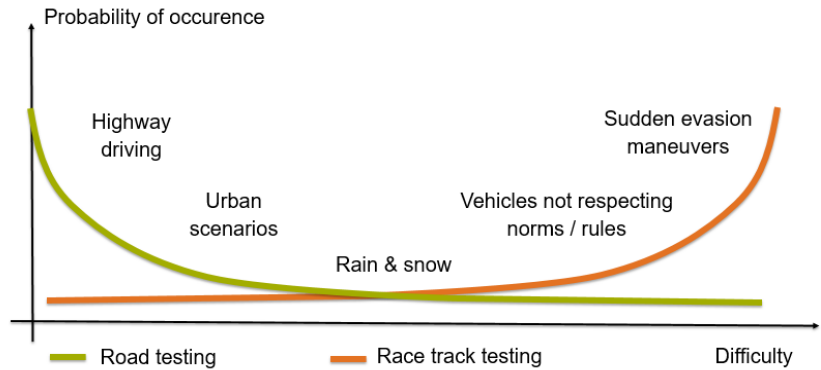

**Figure 2**: Qualitative visualization of the probability to encounter different situations during standard road testing and autonomous racing

The reason for these difficult testing requirements is the statistical distribution of the scenarios autonomous vehicles face. Figure 2 visualizes this distribution as the probability of occurrence for different situations. The main issue for the development process is that the probability of occurrence drops significantly with the increasing difficulty of the scenario. Therefore, it is essential to carefully design a combination of on-road traffic testing and specifically chosen scenarios to allow for a cost- and time-efficient development strategy.

Wheel-to-wheel racing provides an environment where we find a different distribution of scenarios, and therefore, we can focus the development efforts on other aspects. The first category of scenarios that occur much more frequently are multi-vehicle situations with only a very loose set of rules. While it is not allowed to force an opponent off the track or challenge him with erratic maneuvers, there is no specific set of rules comparable to the lane markings and traffic lights we see in typical urban driving situations. These unstructured situations ask for a better understanding and reasoning to successfully predict the opponent's behavior. It has to be derived from its past motion only rather than the structure provided by the scenario. The solutions to these scenarios can provide insights into the solutions for comparable unstructured situations in urban traffic, e.g., construction sites, the appearance of emergency response vehicles, or in the rare circumstance that other traffic participants do not obey the traffic rules.

The second category of scenarios which occur much more frequently in autonomous vehicle racing compared to standard traffic scenarios are sudden evasion scenarios at the limits of vehicle dynamics. In motorsports the algorithms are faced with these challenges on a turn-by-turn basis, trying to exploit the maximum of the tire safely while respecting the track boundaries and opponents. The vast variety of these situations on the racetrack forces the software developers and engineers to design for robust and generalized solutions rather than solving a well (and narrowly) specified set of situations derived from highway or urban driving situations. At the same time, the race track limits the operational design domain (ODD) to essential aspects: Usually races are held under predictable and good weather conditions, vulnerable road users such as cyclists or pedestrians do not have to be considered, and the simple track layout on which the race will be run is known in advance. However, the limited ODD is important to provide a good entry point for research teams and to enable them to focus on the core aspects of research, such as evasive maneuvers at the dynamic limits of driving.

Following these arguments, it becomes clear that autonomous vehicle racing has a different statistical distribution of challenges and can therefore complement the development efforts put into standard traffic scenarios and carefully crafted test scenarios on dedicated testing facilities. In addition, the competitive environment, as well as the inherent safety of motorsport circuits at all times, encourages teams to push the limits of technology in a fast-paced manner.

## 2.2 Indy Autonomous Challenge

The Indy Autonomous Challenge (IAC) aims to leverage the strengths described in the previous section and facilitate the next step in autonomous vehicle technology. Following in the footsteps of its predecessors, the DARPA Grand and Urban Challenges in 2004, 2005 and 2007, it asks university teams to participate in a competitive environment and showcase their research and ideas to the general public. Teams are asked to develop the autonomous racing software for the standardized racing vehicle Dallara AV-21 (see Figure 3), a modified Indy Lights vehicle, and race each other using a ruleset comparable to human drivers on the Indianapolis Motor Speedway. This 4km oval track is mostly known for being the venue for the Indy500 race. Teams are not allowed to modify the hardware of the vehicle and therefore the competition is focused on the software development. More than 30 teams from international universities took the invitation and signed up for the IAC in early 2020 with a concept on how to win the first place and the 1.000.000$ price.

The Indy Lights vehicle is retrofitted with a set of sensors and actuators: four RADAR sensors, six cameras and three LIDAR sensors. Each of the sensor modalities covers a 360-degree field-of-view around the vehicle. This approach enables teams to choose their preferred perception sensors or even implement a redundant system to increase reliability. Furthermore, the vehicle is equipped with a high-precision DGPS receiver. The chassis, as well as the steering, powertrain and brake-system, are similar to the base-vehicle. This allows the vehicle to reach the same high speeds, up to 300 kph, and therefore maintain one of the key characteristics of oval race circuits. The same holds true for the aerodynamic package of the vehicle even though minor modifications had to be done. The main computation platform available to the teams is an x64-based Intel Xeon with 8 CPU-Cores, 32GB RAM and a NVIDIA Quadro GPU in a rugged case to withstand vibrations and mechanical stress in the racecar.

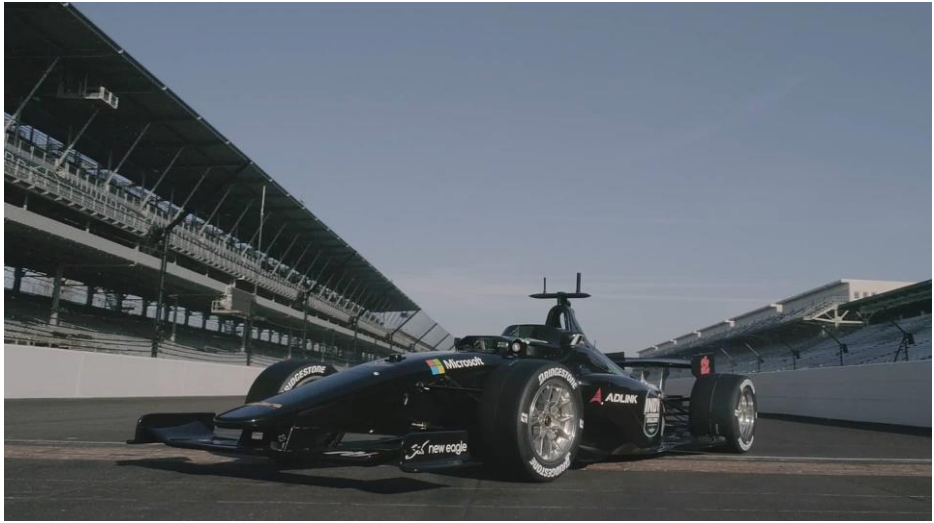

**Figure 3:** The autonomous Dallara - AV21 racecar for the Indy Autonomous Challenge [36]

The competition itself is organized into a split between simulation and real-world testing and competitions. Teams benchmark their code in multiple simulation races with increasing difficulty, from performing a single-vehicle fast lap until the final simulation race in May 2021 where up to eight vehicles compete against each other in multiple heats. From June 2021 ten teams are going to have access to the real Dallara AV-21 vehicles to prepare and test their software on the Indianapolis Motor Speedway and its smaller equivalent, the Lucas Oil Raceway, in Indianapolis.

### 2.3 Software development workflow

Even though the racing circuit provides a safe environment, software errors can lead to situations that cause severe mechanical damage to the race vehicles which results in cost- and time-intense repairs. Even minor errors in assumptions about the vehicle's behavior or its physical capabilities might lead to a spin-out and a crash with the track barriers afterward [37]. For this reason the development workflow of the teams will be one of the main factors for winning the competition. For building a fast autonomous software stack teams need the ability to release new features frequently as well as detecting software issues early in the development process. This is especially challenging due to the size of the TUM Autonomous Motorsport team, formed by a group of 15 PhD candidates as well as more than 40 Bachelor- and Master students.

We employ a strategy based on two pillars to tackle challenges on an organizational level: An agile software development process organized on the collaboration platform GitLab as well as the utilization of a Hardware-in-the-Loop simulator for up to eight autonomous agents for development purposes (see Figure 4 for an exemplary architecture diagram with three vehicles). The latter is built around a Real-Time platform (Speedgoat Performance Real-Time Target Machine) for vehicle dynamics simulation and a GPU-Server (Intel Xeon with 2x20 CPU Cores and 2x NVIDIA RTX 3080) for the environment simulation. The autonomous vehicle computers are high-performance workstations to resemble the vehicle computer from a resource's point of view. Due to resource constraints, the environment perception can only simulate one set of perception sensors for one autonomous vehicle in real-time. The other seven vehicles are provided with an idealized perception similar to Vehicle-to-Vehicle communication. This allows to deploy a realistic Hardware-in-the-Loop environment for one of the vehicles with true multi-agent simulation. The vehicle dynamics simulation is built around a sophisticated nonlinear dual-track model with a Pacejka combined tire model [38]. Furthermore, sensors and actuators are modeled with their specific response and noise characteristics and external effects such as wind or other random forces acting upon the chassis are simulated. This challenging environment prevents overfitting to idealized simulation data and guarantees the fast transition from the virtual world to the real vehicle.

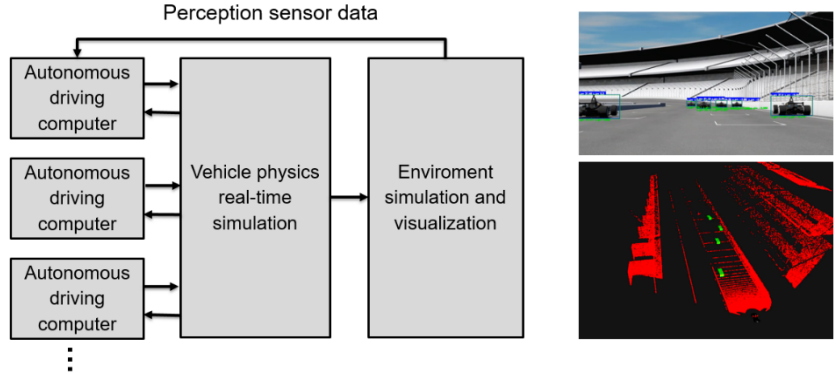

**Figure 4:** High-level overview of the Hardware-in-the-Loop simulation environment for three autonomous vehicles. It can be extended for up to eight autonomous vehicles (left). Camera and LIDAR data generated by the perception sensor simulation (right).

## 3    Performance aspects in autonomous vehicle racing

The oval track at Indianapolis as well as the strong focus on multi-vehicle scenarios within the competition lead to a focus on two things in the development process: First and foremost it is important to maximize the achieved top speed of the autonomous vehicle while being able to safely navigate dynamic multi-vehicle scenarios. The nearly one kilometer long straights allows the ego vehicle to overtake an opponent already at relatively low speed differences around 5-10 kph. The key aspect here is that all software parts (perception, planning and control) need to be able to operate at the maximum vehicle speed of 300 kph. We therefore want to point out that it is necessary to approach this task with a holistic perspective on all parts of the software stack. This is a strong contrast to achieving a single fastest lap in a qualifying scenario which mainly requires the software to accurately control the vehicle at the handling limits and does not require complex perception or planning algorithms. The rest of this chapter will give a short introduction to the software architecture and algorithms chosen to solve this challenge as well as the major difficulties encountered during the development work.

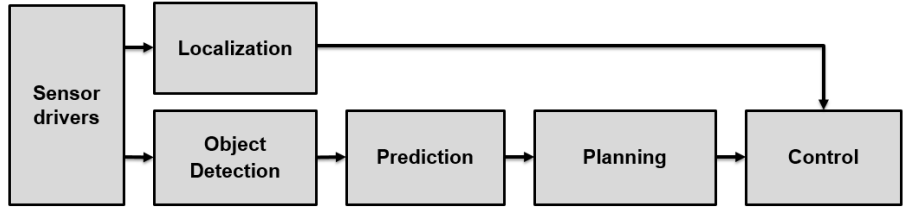

**Figure 5:** Software architecture of the TUM Autonomous Motorsport software for the Indy Autonomous Challenge

### 3.1 Software architecture

This paper will give a coarse overview of the software architecture we propose (Figure 5) to solve the autonomous multi-vehicle racing task posed by the IAC. It is built upon the common strategy to decompose the complex driving into multiple subtasks of localization, object detection, opponent prediction, behavior planning and motion control.

The localization is done by fusing a LIDAR-based localization based on the distances to the track bounds as well as the DGPS. The former is mainly used for lateral localization while the latter gives the longitudinal position along the track. The object detection is designed to be fully redundant: It uses the object list generated by the RADAR sensor, a camera detection based on bounding box estimation and a known-height transformation and two LIDAR detection pipelines. One of them is based on a Deep-Learning approach while the other uses conventional clustering techniques. This variety of sensor modalities and algorithms allows us to build a fully redundant system.

The behavior prediction is built around two strategies: A short-term prediction based on a simple vehicle physics model generates reliable estimates for the behavior of the next 1-2 seconds. It is accompanied by a data-based prediction algorithm which uses previously seen scenarios to predict the motion of the opponent vehicles for the next five seconds. This gives the planning algorithm an estimate of what other vehicles might do in the future. The planning itself utilizes a spatio-temporal graph-based search approach to generate a coarse target trajectory, which is then re-optimized and tracked by a model-predictive controller.

### 3.2 High sensor detection distances

The first aspect we want to cover in this analysis is the impact of the maximum object detection range on the achievable autonomous racing performance. The limiting performance factor here is the maximum difference speed at which the ego vehicle is capable of executing a safe evasion maneuver. Higher maximum difference speed also leads to higher overall top speed. In a worst-case scenario, the vehicle might encounter an opponent driving slowly or even coming to a standstill during the race due to a component or software failure. While it is easily possible to calculate the required braking distance based on the difference speed and maximum deceleration, this is more complex when the vehicle attempts an evasive maneuver rather than an emergency brake. We therefore conducted a case study on our Hardware-in-the-Loop simulation environment with the following setup: the ego-vehicle drives a single fast lap and encounters an obstacle at standstill on the start-finish straight (right-side of Figure 6). The influence of the sensor detection range is evaluated at different speed levels. Higher speeds are expected to require larger detection distances to perform a successful evasion maneuver. The experiments have been carried out with the full prediction, planning and control software modules but an idealized perception providing exact object positions. The detection ranges for which a successful evasion maneuver was performed are depicted on the left in Figure 6. In comparison, we show the required theoretical braking distance at maximum acceleration for the speed chosen in the experiment. It is noteworthy that the maximum required detection distance of approximately 100 m does not increase significantly anymore above 200 kph. The reason for this is the fact that the motion planning does not trigger an evasive maneuver before the vehicle actually reaches this distance. Furthermore, we want to emphasize that the real detection distance required will be slightly higher due to the computational delay introduced by the object detection algorithms. For a speed of 300 kph and a processing time of 200 ms, the required de-tection distance would increase approximately by 17 m (driven distance within the processing time at the given speed). Overall, we laid out a target of 130-150 m detection distance for the development of the object detection algorithms.

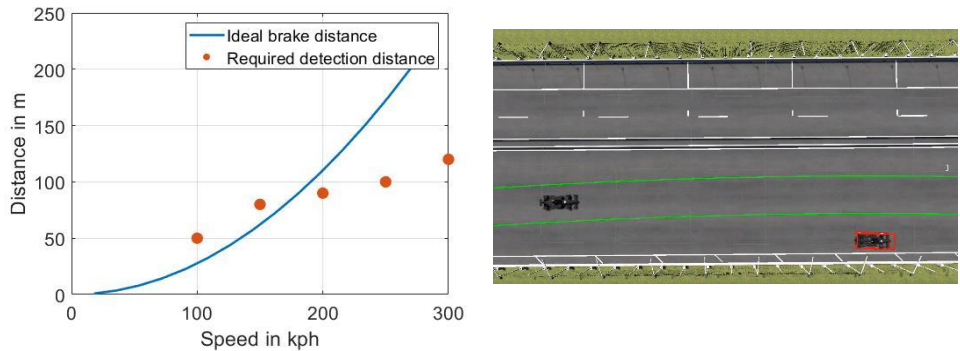

**Figure 6:** The required detection distance for a reliable evasion maneuver is depicted on the left. The ideal brake distance values are given for a maximum deceleration of 14mps$^2$. The scenario used to for analysis is depicted on right. The red box indicates the detected opponent vehicle at standstill. The green lines limit the admissible driving corridor for the control software. The simulations were conducted with an idealized object detection and the full prediction, planning and control software.

### 3.3 Fast reaction times to master multi-vehicle dynamic scenarios

Despite the challenge to achieve reliable and robust vehicle dynamics control, autonomous wheel-to-wheel racing requires a thorough understanding of the situation around the ego-vehicle as well as a reasonably good prediction of what is going to happen. Strict rules, as in road traffic, are replaced by competing race cars with conflicting goals, namely winning the race. Therefore, opponents may switch between different distinct strategies depending on their own and their competitor's situation. It must also be noted that not only the prediction of the opposing vehicles influences the planning of one's own trajectory, but also vice versa the planned trajectory affects the opposing vehicles and thus their prediction. Accordingly, the behavior of opposing vehicles can never be predicted with absolute certainty. One prominent example might be the case that the ego-vehicle is chasing two opponents fighting for a position in front while approaching one of the high-speed turns (Figure 7). Even the most experienced driver will not be able to predict with absolute certainty whether the vehicle trying to overtake will successfully complete its maneuver or it needs to abort and keep its position. There are now two approaches how to handle this situation in the motion planning algorithm:

In the more sophisticated approach, the prediction algorithm would output a detailed stochastic model of all possible outcomes for the situation based on e.g. a large database of scenarios. It should be noted here, that this stochastic model would be required to be multi-modal, i.e. assigning probabilities to distinct results such as an successful overtake as well as an aborted overtake. Given this information, the planning algorithm could optimize for a time-optimal trajectory with respect to certain risk constraints.

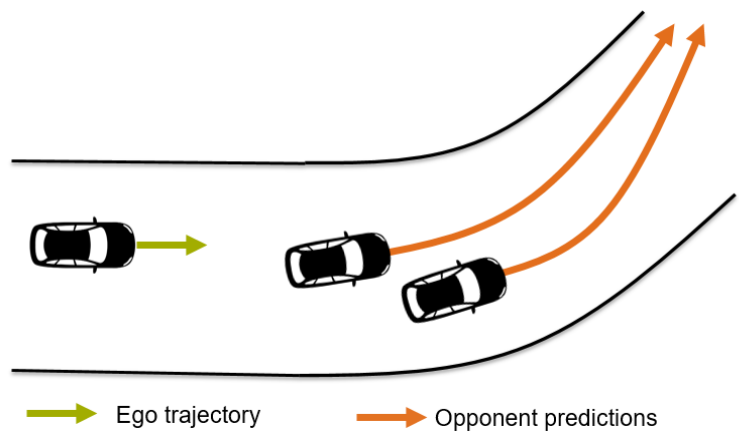

    → Ego trajectory      → Opponent predictions

**Figure 7:** Multi-vehicle scenario with two opponents fighting for a position in front of the ego vehicle

However, this approach has two disadvantages: First, it requires a huge scenario data-base which is difficult to build and maintain. This is caused by the fact that we would need autonomous vehicles being able to race each other to build the database before actually having access to the data to fill the database. Second, the optimization over multi-modal stochastic distribution is a computationally challenging task even on state-of-the-art compute-platforms and not feasible within fractions of a second.

In contrast to this sophisticated strategy, we propose to utilize a more computation efficient approach. Rather than building a complex stochastic model of the situation that is prone to false assumptions and difficult to scale for more than a few vehicles, we rely on the advantages of frequent updates and feedback. With this concept, the prediction outputs only the most likely behavior of the opponent vehicles and therefore implicitly decides for one of the multi-modal outcomes from the previous concept. The planning can utilize this decision and solve a deterministic optimization problem under the assumption that the predictions are correct. It is obvious that this assumption is likely to be not fulfilled in real-world driving. However, this deficiency can be mitigated by frequent updates of the prediction and the planning. This is a characteristic known from human drivers as well: Fast reaction times lead to safer and more reliable driving styles. This comparison also helps to estimate a minimum requirement for the update rate required to master difficult scenarios: Human reaction times lie within the range of 0.5-1.0 seconds for average drivers [39]. Racing drivers are likely to show superior performance, which leads us to set a target of 0.3-0.5 seconds to achieve comparable performance to a human race driver. Note that this time requirement includes all parts of the autonomous driving pipeline depicted in Figure 5 and therefore poses a huge challenge in algorithm selection and implementation.

### 3.4    Reliable motion control in the presence of uncertainty

One of the challenges in developing a reliable control system for an autonomous racing vehicle is the complex modeling of the dynamic behavior of the vehicle. There are several sources of uncertainty, e.g. the actuator dynamics and the tire model. The latter are usually represented via empiric models (such as the Pacejka model [38]) from extensive test-rig measurements or specific vehicle dynamics tests. However, this approach is prone to suggest a false sense of accuracy for the event of changing environmental conditions. To name only two examples, the road surface on different tracks as well as a change in tire temperature during the race are likely to affect the accuracy of these models significantly. In the spirit of Occam's razor and its principle of parsimony, we therefore propose to focus on rather simple friction limited point-mass models in the area of motion planning and control [6, 7]. In addition, we specifically consider the remaining inaccuracies in the vehicle dynamics by a Robust Model Predictive Control scheme on the motion control level and make use of fast low-level feedback loops for the lateral and longitudinal vehicle dynamics. This approach significantly reduces the required amount of last-minute parameter tuning in case of changing track conditions.

# 4    Conclusion & Future work

This paper gives an overview of the TUM Autonomous Motorsport team activities and the reasoning to enter the Indy Autonomous Challenge, the first wheel-to-wheel race with full-scale autonomous racecars. The key advantage with respect to standard urban scenario testing is a much higher likelihood of unstructured multi-vehicle situations as well as maneuvers at the handling limits. This supports innovation in these areas and will help solving the difficult edge-cases in autonomous driving.

In the second part of the paper, we gave insights on our development priorities by analyzing three key performance indicators for autonomous vehicle racing: Robust motion control at the handling limits, fast reaction times in multi-vehicle scenarios as well as high sensor detection range.

Our current focus lies on adopting the developed software stack to the Dallara AV-21 racing vehicle and preparation of the first test sessions in June 2021. We are going to validate our models and algorithms on the Indianapolis Motor Speedway and expect further insights into the interplay of the algorithms under real-world conditions. Furthermore, several of the algorithms used are already or will be published soon on our GitHub page [40].

## Contributions

Alexander Wischnewski is the main author of this paper and contributed essentially to the presented discussions and experiments. Maximilian Geisslinger, Johannes Betz, Tobias Betz, Felix Fent, Alexander Heilmeier, Leonhard Hermansdorfer, Thomas Herrmann, Sebastian Huch, Phillip Karle, Felix Nobis, Levent Ögretmen, Matthias Rowold, Florian Sauerbeck, Tim Stahl, Rainer Trauth, Markus Lienkamp and Boris Lohmann are all team members of the TUM Autonomous Motorsport team and contributed equally to different parts of the software stack, the presented discussions as well as the overall design of the research project.